\documentclass[11pt,a4paper]{article}
\usepackage{times,latexsym}
\usepackage{url}
\usepackage[T1]{fontenc}

\usepackage[acceptedWithA]{tacl2021v1}

\usepackage{xspace,mfirstuc,tabulary}

\usepackage{graphicx}
\usepackage{tabularx} 
\usepackage{soul}
\usepackage{multirow}

\usepackage{epstopdf}
\usepackage[utf8]{inputenc}

\usepackage{hyperref}
\usepackage{xstring}

\usepackage{arydshln}
\usepackage{tikz}
\usepackage{enumitem}

\usepackage{soul}

\usepackage{color}

\usepackage{microtype}
\usepackage[most]{tcolorbox}
\usepackage{pifont}
\usepackage{scrextend}
\usepackage{pifont}
\usepackage{tabularx}
\usepackage{booktabs}
\usepackage{graphicx}
\usepackage{subcaption}
\usepackage{xurl}
\usepackage{stmaryrd}
\usepackage{amsmath}
\usepackage{cleveref}
\usepackage{xcolor}
\usepackage{tikz-dependency}
\usepackage{flowchart}
\usepackage{pgf}
\usetikzlibrary{arrows.meta,shapes.symbols,shapes.geometric,shadows}
\usepackage{varwidth}
\usepackage{tabularx}
\usepackage{tablefootnote}
\usepackage{makecell}
\usepackage{multirow}
\usepackage{array}
\usepackage{arydshln}
\usepackage{lingmacros}
\usepackage{enumitem}
\usepackage{soul}
\setul{}{0.2ex}
\setulcolor{black}

\newif\iftaclinstructions
\taclinstructionsfalse 
\iftaclinstructions
\renewcommand{\confidential}{}
\renewcommand{\anonsubtext}{(No author info supplied here, for consistency with
TACL-submission anonymization requirements)}
\newcommand{\instr}
\fi

\iftaclpubformat 

\else

\fi

\usepackage[acronym, nomain]{glossaries} 
\glsdisablehyper 
\newacronym{ai}{AI}{Artificial Intelligence}
\newacronym{nlp}{NLP}{natural language processing}
\newacronym{cl}{CL}{computational linguistics}
\newacronym{gui}{GUI}{graphical user interface}
\newacronym{ner}{NER}{named entity recognition}
\newacronym{cv}{CV}{computer vision}
\newacronym{dnn}{DNN}{deep neural networks}
\newacronym{srl}{SRL}{Semantic Role Labeling}
\newacronym{crl}{[INN]}{I Need a Name}
\newacronym{esl}{ESL}{Event Structure Lexicon}
\newacronym{r2vq}{R2VQ}{Recipe-to-Video Questions}
\newacronym{vqa}{VQA}{Visual Question Answering}
\newacronym{qa}{QA}{Question Answering}
\newacronym{qg}{QG}{Question Generation}
\newacronym{qf}{QF}{Question Family}
\newacronym{cb}{CB}{Competence-based}
\newacronym{dp}{DP}{Dense Paraphrasing}
\newacronym{dped}{DP'ed}{Dense Paraphrased}
\newacronym{dps}{DPs}{Dense Paraphrases}
\newacronym{hrp}{HRP}{Human Readable Paraphrase}
\newacronym{mrp}{MRP}{Machine Readable Paraphrase}
\newacronym{iaa}{IAA}{Inter-Annotator Agreement}
\newacronym{deep}{AAE}{Anonymized Annotation Environment}
\usepackage{amsthm}
\theoremstyle{definition}
\newtheorem{thm}{Theorem}[section]
\newtheorem{defn}[thm]{Definition} 

\title{Dense Paraphrasing for Textual Enrichment}

\author{
  Jingxuan Tu, Kyeongmin Rim, Eben Holderness, James Pustejovsky \\
  Department of Computer Science \\
  Brandeis University  \\
  Waltham, Massachusetts  \\
  \texttt{\{jxtu,krim,egh,jamesp\}@brandeis.edu}
}

\date{}

\begin{document}
\maketitle
\begin{abstract}
Understanding inferences and answering questions from text requires more than merely recovering surface arguments, adjuncts, or strings associated with the query terms. As humans, we interpret sentences as contextualized components of a narrative or discourse, by both filling in missing information, and reasoning about event consequences. In this paper, we define the process of rewriting a textual expression (lexeme or phrase) such that it reduces ambiguity while also making explicit the underlying semantics that is not (necessarily) expressed in the economy of sentence structure as {\it \gls{dp}}.
We build the first complete \gls{dp} dataset, provide the scope and design of the annotation task, and present results demonstrating how this \gls{dp} process can enrich a source text to improve inferencing and \gls{qa} task performance. The data and the source code will be publicly available.

\end{abstract}

\section{Introduction}
\label{sec:intro}
Two of the most important components of understanding natural language involve
recognizing that many different textual expressions can correspond to the same meaning, and 
detecting those aspects of meaning that are not 
present in the surface form of an utterance or narrative.
Together, these involve broadly three kinds of interpretive processes: 
(i) recognizing the diverse variability in linguistic forms that can be associated with the same underlying semantic representation (paraphrases); (ii) 	 (e.g., {\it stir vigorously}); and (iii) interpreting or computing the dynamic
consequences of actions and events in the text (e.g., {\it slicing an onion}
brings about {\it onion slices}).

The first of these, the problem of paraphrasing, has been addressed computationally since the early days of \gls{nlp}.
The second and third dimensions of sentence meaning mentioned above, however, are more difficult  to model with current ML approaches, which rely heavily on explicit textual strings to model semantic associations between the elements in the input. 
Many question answering systems, for example, rely on such syntagmatic forms in the training data for modeling potential associations that contribute to completion or generation task performance. Hence, if predicates or arguments are missing, implied, or interpreted from context, there is nothing to encode, and consequently little to decode as output, as well. Consider the following example from the traditional paraphrasing task. The text difference between the input and output only comes from a lexical substitution, rather than the rephrasing or addition of hidden arguments. 

\enumsentence{
{ Paraphrasing:\\
Chop onions, saute until browned. $\xrightarrow{}$\\
\textit{\textbf{Cut} onions, saute until \textbf{done}.}\\
\label{inline-paraphrasing}}}

To solve this problem, some recent attempts have been made to enrich surface sentence forms that are missing information through ``decontextualization'' procedures that textually supply  information which would make the sentence interpretable out of its local context  \cite{choi2021decontextualization,elazar2021text,wu2021conqrr}. Choi \shortcite{choi2021decontextualization} formally defines decontextualization as:

\begin{defn}
    {\bf Decontextualization}:
    Given a sentence-context pair $(S,C)$, a sentence $S'$ is a valid decontextualization of s if: (1) the sentence $S'$ is interpretable in the empty context; and (2) the truth-conditional meaning of $S'$ in the empty context is the same as the truth-conditional meaning of $S$ in context $C$. 
\end{defn}

The decontextualization task focuses on enriching text through anaphora resolution and knowledge base augmentation, which works well on arguments or concepts that can be linked back to existing knowledge sources, such as Wikipedia. Consider the following example of the decontextualization task. It is able to decontextualize \textit{Barilla sauce} in (\ref{inline-decont}a), but does not reintroduce any semantically hidden arguments from the context in (\ref{inline-decont}b), making inferences over such sentences difficult or impossible. 

\enumsentence{
{\it Decontextualization}:\\
a. Add Barilla sauce, salt and red pepper flakes. $\xrightarrow{}$\\
\textit{Add Barilla sauce, \textbf{the tomato sauce,} salt and red pepper flakes.} \\
b. Simmer 2 minutes over medium heat. $\xrightarrow{}$\\
\textit{ Simmer 2 minutes over medium heat.}
\label{inline-decont}}






In this paper, we argue that the problems of paraphrasing and decontextualizing are closely related, and part of a richer process of what we call {\it Dense Paraphrasing}. This combines the textual variability of an expression's meaning (paraphrase) with the amplification or enrichment of meaning associated with an expression (decontextualization). 

While a paraphrase is typically defined as a relation between two expressions that convey the same meaning \cite{bhagat2013paraphrase}, it has also been used to clarify meaning through verbal, nominal, or structural restatements that preserve (and enhance) meaning \cite{smaby1971paraphrase,kahane1984meaning,mel1995phrasemes,mel2012phraseology},   in particular the notion of ``entailed paraphrase'' \cite{culicover1968paraphrase}: ({\it author}, {\it person who writes}), ({\it sicken}, {\it to make ill}),  ({\it strong}, {\it potent (of tea)}). 

This is clearly related to  recent efforts at decontextualizing linguistic expressions with ``contextual enrichments" \cite{choi2021decontextualization}. What these approaches do not focus on, however, is the notion of enrichment of the expression through both its lexical semantics and its dynamic contribution to the text in the narrative. 
We define a Dense Paraphrase (DP) as follows:

 \begin{table*}[h]
    \centering
    \begin{tabular}{p{0.2\linewidth} p{0.75\linewidth}}
      \multicolumn{2}{p{0.95\linewidth}}{\emph{Passage:} \textcolor{orange}{Peel} and \textcolor{brown}{cut apples into wedges}. 
      \textcolor{red}{Press apple wedges partly into batter}. \textcolor{cyan}{Combine  sugar and cinnamon}. \textcolor{violet}{Sprinkle over apple}. \textcolor{green}{Bake at 425 degF for 25 to 30 minutes.}} \\ \hline
      \multicolumn{2}{p{0.95\linewidth}}{\emph{\gls{dped} Passage:}} \\ 
      
    \multicolumn{2}{p{0.95\linewidth}}{\textcolor{orange}{Using peeler, peel \ul{apples}, resulting in \ul{peeled apples}}; and \textcolor{brown}{using knife on cutting board, cut \ul{peeled apples} into \ul{peeled wedges}}.} \\
    \multicolumn{2}{p{0.95\linewidth}}{\textcolor{red}{Using hands, press \ul{peeled apple wedges} partly into batter in the cake pan}.}\\ \multicolumn{2}{p{0.95\linewidth}}{\textcolor{cyan}{Combine sugar and cinnamon in a bowl, resulting in cinnamon sugar}.}\\
    \multicolumn{2}{p{0.95\linewidth}}{\textcolor{violet}{Sprinkle cinnamon sugar over \ul{peeled apple wedges} in batter in cake pan, resulting in \ul{appelkoek}}.}\\
    \multicolumn{2}{p{0.95\linewidth}}{\textcolor{green}{In oven, bake \ul{appelkoek} at 425 degF for 25 to 30 minutes, resulting in  \ul{baked appelkoek}.}} \\ 

    \end{tabular}
    \caption{Example recipe passage. Color-coded text spans represent locations of cooking events in the input text where \gls{dps} are generated to enrich local context. Underlined text shows a chain of coreferential entities for the ingredient ``apple''.  } 
    \label{tab:question_example}
\end{table*}
 
\begin{defn}
    {\bf \glsentrylong{dp}}: 
    Given the pair, $(S,P)$, where $S$ is a source expression, and $P$ is an expression, we say $P$ is a valid {\it Dense Paraphrase} of $S$ if: $P$ is an expression (lexeme, phrase, sentence) that eliminates any contextual ambiguity that may be present in $S$, but that also makes explicit any underlying semantics that is not otherwise expressed in the economy of sentence structure, e.g., default or hidden arguments, dropped objects or adjuncts. $P$ is both meaning preserving (consistent) and ampliative (informative) with respect to $S$. 
\end{defn}

The following shows the \gls{dps} of the sentences from  examples (\ref{inline-paraphrasing}) and (\ref{inline-decont}). Compared to the aforementioned tasks, a \gls{dp} aims to recover semantically hidden arguments through: (1) a broader view of the context of the text; and  (2) commonsense or best educated guesses from humans (i.e., text spans with \underline{underlines} from the example).

\enumsentence{
{ 
Chop onions, saute until browned. $\xrightarrow{}$\\
\textit{Chop onions \underline{\textbf{on a cutting board}} \underline{\textbf{with a knife}} to get \textbf{chopped onions}, saute \textbf{chopped onions} \underline{\textbf{on a pan}} \underline{\textbf{with a spatula}} until browned, resulting in \textbf{sauted chopped onions}.}\\
------------------------------------------------------\\
Add Barilla sauce, salt and red pepper flakes to the saucepan. Simmer 2 minutes over medium heat. $\xrightarrow{}$\\
\textit{Add Barilla sauce, salt and red pepper flakes to the saucepan \textbf{\underline{by hand} to get sauce mixture}. Simmer the \textbf{sauce mixture} 2 minutes \textbf{in the saucepan} over medium heat to get \textbf{simmered sauce mixture}.}\\
\label{inline-dp}}}

We argue that our work can potentially help and complement these generation tasks by enriching the source text with information that is not on the surface, by  either additional text strings or vector representations. 
To show the usage of \gls{dp}, we evaluate our method through \gls{qa} tasks on dense-paraphrased questions.

In the remainder of the paper, we first review related work and background (\textsection\ref{sec:background}), and give more detailed definitions of the \gls{dp} schema (\textsection\ref{sec:method}).
We then introduce a dataset we have created to support our implementation of the \gls{dp} operation (\textsection\ref{sec:dataset-intro}), immediately followed by the details of how we collected and annotated this dataset (\textsection\ref{sec:data-annotation}). \textsection\ref{sec:experiments} provides details of experiments we conducted to validate the utility of the proposed methodology, along with their results. Then conclude our work in the final (\textsection\ref{sec:conclusion}).

\section{Background}
\label{sec:background}

There is a long history in linguistics, dating back to the early 1960s, of modeling linguistic syntagmatic surface form variation in terms of transformations or sets of constructional variants \cite{harris1954distributional,harris1957co}
\cite{hiz1964role}. 
When these transformations are viewed ``derivationally", i.e., as an ordered application of rules over an underlying form, the resulting theory is in the family of generative grammars \cite{chomsky1957,bach1964introduction}. If they are seen as undifferentiated  choices over surface constructional forms of an expression, the resulting theory can be called a paraphrase grammar \cite{hiz1964role,smaby1971paraphrase,culicover1968paraphrase}.  Formally, a paraphrase is a relation between two lexical, phrasal, or sentential expressions, $E_i$ and $E_j$, where meaning is preserved \cite{smaby1971paraphrase}. 

For \gls{nlp} uses, paraphrasing has been a major part of machine translation and summarization system performance  \cite{culicover1968paraphrase,goldman1977sentence,muraki1982semantic,boyer1985generating,mckeown1983paraphrasing,barzilay1999using,bhagat2013paraphrase}. 
In fact, statistical and neural paraphrasing is a robust and richly evaluated component of many benchmarked tasks, notably MT and summarization \cite{weston2021generative}, as well as Question Answering \cite{fader2013paraphrase} and semantic parsing \cite{berant2014semantic}.  To this end, significant efforts have gone towards the collection and compilation of paraphrase datasets for training and evaluation  
\cite{dolan2005automatically,ganitkevitch2013ppdb,ganitkevitch2014multilingual,pavlick2015ppdb,williams2017broad}. 

In addition to above meaning-preserving paraphrase strategies, there are several directions currently that use strategies of ``decontextualization'' or ``enrichment'' of a textual sequence, whereby  missing, elliptical, or underspecified material is re-inserted into the expression. The original and target sentences are compared and judged by an evaluation as a text generation or completion task \cite{choi2021decontextualization,elazar2021text}.


Enrichment of VerbNet predicates can be seen as an early attempt to provide a kind of Dense Paraphrasing for the verb's meaning. 
In \newcite{im2009annotating,im2010annotating}, the basic logic of \textit{Generative Lexicon}'s subevent structure was applied to  VerbNet classes, to enrich the event representation for inference. The VerbNet classes were associated with event frames within an {\it \gls{esl} } \cite{im2010annotating}, encoding the subevent structure of the predicate. If the textual form for the verb is replaced with the subeventual description itself, classes such as  {\small\texttt{change\_of\_location}} and {\small\texttt{change\_of\_possession}} can help encode and describe event dynamics in the text, as shown in  \cite{brown2018integrating,dhole2021synqg,brown2022semantic}. For example, the VerbNet entry {\it drive} is enriched with the ESL subevent structure below:

\enumsentence{
\hspace*{1mm} {\small \textbf{drive} in \textit{John drove to Boston}\\
\hspace*{5mm}se1: pre-state: not\_located\_in (john,boston)\\
\hspace*{5mm}se2: process: driving (john)\\
\hspace*{5mm}se3: post-state: located\_in (john,boston)\\
\label{drive}}}

\noindent
 In the remainder of the paper, such techniques  will be utilized   as part of our Dense Paraphrasing strategy to enrich the surface text available for language modeling algorithms.

\section{Method: \glsentrylong{dp}} 
\label{sec:method}

In this section, we detail the procedure involved in creating \gls{dps} for a text. 
Compared to  decontextualization, \gls{dp} can be seen as similar, but is a much broader method for creating sets of semantically equivalent or ``enriched consistent" expressions, that can be exploited for either human or machine consumption. 


Unlike traditional paraphrases that are evaluated in terms of how faithful and complete they are, while preserving the literal interpretation of the source, the goal of our task is to generate \emph{dense} paraphrases that can be merged with the concept of semantic enrichment, to give rise to a set of paraphrases of semantically enriched and decontextualized expressions. 
%
%
%
We distinguish between two contexts of use for a paraphrase:

\begin{defn}
{\bf \gls{hrp}}: the redescription of the source expression, $s$, generated as a paraphrase of $s$, $s^p$, is intended to be read, viewed, or heard by a human audience. Context, style, genre, register, and voice may dictate nuanced variations in the resulting form of the paraphrase;
\end{defn}
\begin{defn}
{\bf \gls{mrp}}: the source expression, $s$, is enriched with descriptive content and contextualized information that turns implicit content into  explicit textual expressions. The output of \gls{mrp} is logically consumed by a downstream model, such as a question-answering system, that can utilize the richer local environment for improved accuracy on a variety of reasoning tasks. 
\end{defn}

Consider the following example of the \gls{dp} of the original recipe sentence.  Table \ref{tab:question_example} shows a dense paraphrased passage from this data. 
The original text and corresponding dense paraphrased text are associated with the same color. Both \gls{hrp} and \gls{mrp} formsd from the same sentence are illustrated. The information that is encoded in both paraphrases is identical. \gls{hrp} includes the insertion of additional prepositions and the proper ordering of textual components, while \gls{mrp} includes metadata content to structure the arguments and relations.

\enumsentence{
{Chop onions, ... $\xrightarrow{}$\\
\gls{hrp}: \textit{Chop onions on a cutting board with a knife
to get chopped onions}\\
\gls{mrp}: {\small{Chop \{\texttt{TOOL}:knife \# \texttt{HABITAT}:cutting board \# \texttt{OUTCOME}:chopped onions\} onions \{\texttt{INGRE\_OF}:chop\}}}\\
\label{inline-dp2}}}


 
\subsection{\glsentrylong{dp} Procedure}
\label{ssec:procedure}

In this section, we describe the mechanisms involved in creating a {\it \gls{dp}} from a source text. 
In this work we will focus on the \gls{mrp}, since our present goal is creating \gls{dps} that can be used in the service of \gls{nlp} applications. 
Specifically, we adopt a template-based method along with heuristics to: (1) generate dense-paraphrases that account for hidden entities and entity subevent structure; and (2) convert them to quasi-grammatical text formats for machine consumption. 

 We provide the source narrative with a dynamic {\it \gls{dp}} of the surface text, which both decontextualizes the expression \cite{choi2021decontextualization,elazar2021text}, but also enriches the textual description of both events and participants to reflect the changes in the object due to the events. \gls{dp} involves identifying conventional coreference chains where the entities from the chain are identical. Importantly, this procedure also 
 includes additional textual descriptions involving: (1) recovering ``hidden''  arguments; and (2) the results of subevent decomposition \cite{pustejovsky1995generative,im2010annotating}, which create the coreference relation between the entity and its hidden or transformed mention elsewhere (e.g. \textit{apples} $\rightarrow$ \textit{apple wedges} $\rightarrow$ \textit{applekoek}).

\subsubsection{Recovering ``hidden'' Arguments}
\label{sssec:saturation}
We define a ``hidden'' argument to a predicate as an event participant that is not  present in the surface form of the text. Given this, we distinguish two subtypes of hidden arguments:
\footnote{\citet{msamr} uses {\it implicit role} to cover drop arguments in AMR. However, shadow arguments do not appear to fall under their category of implicit roles. To our knowledge, this work is the first to annotate the information associated with shadow arguments in verbal constructions.

}
\begin{itemize}
    \item  \textbf{Drop argument}: A drop argument is an argument to a predicate that has been elided or left unexpressed in the syntax. Such elisions occur when the antecedent has been   mentioned in a previous sentence and can be recovered from the context in the document.
    \item \textbf{Shadow argument}: A shadow argument is semantically incorporated in the meaning of the event predicate itself; e.g., an implicit tool or ingredient that is not mentioned but presupposed \cite{pustejovsky1995generative,JohnsonEtAl:01a}.
\end{itemize}

We manually annotate our data to identify all the hidden arguments (both syntactic and semantic) associated with an event predicate. This effectively ``saturates" the lexical frame \cite{fillmore85:_frames} by supplying those frame elements needed to perform richer inferential tasks.

\subsubsection{Recovering  Entity Properties from Event Structure} 
\label{sssec:subevent}

As mentioned above, 
\gls{esl} represents an event as having three parts: {\bf begin (Be)}, {\bf inside (Ie)}, and {\bf  end (Ee)}. In our method, we use this subevent structure to not only track the begin and end state of an event, but create the begin and end {\it textual redescription} of the changed entity itself. To illustrate, consider an example of the \gls{dp} associated with the subevent descriptions for two distinct two-event sequences from the recipe corpus. 
\eenumsentence{\small{
\item {\it Chop$_{e1}$ onions. Saut\'e$_{e2}$ until browned.} $\xrightarrow{}$ \\
{\bf B$_{e1}$}: "unchopped onions"; \\
{\bf E$_{e1}$}: "chopped onions"; \\
{\bf B$_{e2}$}: "unsaut\'eed chopped onions"; \\
{\bf E$_{e2}$}: "saut\'eed chopped onions" \\
\item  {\it Chop$_{e1'}$ onions. Add$_{e2'}$ to a hot pan.} $\xrightarrow{}$ \\
{\bf B$_{e1'}$}: "unchopped onions"; \\
{\bf E$_{e1'}$}: "chopped onions"; \\
{\bf B$_{e2'}$}: LOC("chopped onions","cutting board");\\
{\bf E$_{e2'}$}: LOC("chopped onions","hot pan") \\
}}
\label{onions}


As shown in the examples, we define two kinds of event type and associated end state, namely {\sc transformation} and {\sc location-change} \cite{im2010annotating}. 
The end state {\sc transformation} denotes a change of the object in shape, size or color, etc., and its surface form is changed correspondingly. {\sc location-change} denotes a change of the object location, and this change will not be reflected on the surface form of the object itself, rather a new location is identified as the destination of the object.

In (\ref{onions}a), the \emph{saut\'e} event is of type transformation, resulting in an entity redescription of \emph{saut\'eed chopped onions}. The \emph{add} event in (\ref{onions}b) involves location change that moves the onions from the cutting board to the pan, but does not change the status of the object.

\section{\glsentrylong{dp} Dataset}
\label{sec:dataset-intro}
 We use the text data from the subdomain of cooking recipes to build the \gls{dp} dataset. Compared to texts of news or narratives, procedural text such as recipes and user manuals tend to be task-oriented, and the main content is split into steps that describe small goals to accomplish the final task. We believe such texts are a excellent fit for our task as it involves the understanding of how to reach a goal locally for each step, as well as how each step contributes to the final task globally. Further, the step-wise progression inherent in the goal-oriented narrative contributes both an interpretative dynamics as well as contextualized elision of arguments.
 
We collect a \gls{dp} dataset that consists of 1,000 English cooking recipes (Table \ref{tab:r2vq_stats}) from two open-source recipe wikis.\footnote{\url{https://recipes.fandom.com/} , \url{http://foodista.com/}} The dataset covers about 14K events, 33K entities (15K hidden, 18K explicit), and 25K coreference chains. Each annotated recipe document consists of its provenance, title, and a list of sentences. Each recipe document also contains a set of events, and each event consists of semantic role labeling, both explicit and hidden entities, and relations. The annotated event can be dense-paraphrased into both machine-readable and human-readable format. The dynamics of the events from this same recipe can also be inferred from the coreference chains. The complete dataset is represented in the CoNLL-U format.\footnote{\url{https://universaldependencies.org/format.html}}

\begin{table}[h]
\centering
\resizebox{1\linewidth}{!}{
\begin{tabular}{l|rrr}
           & Train & Dev & Test   \\ \hline
{\# of recipes} &800 &100 &100\\ \hline
{Avg. \# of sentences per recipe} &8 &7.9 &7.8\\
{Max. \# of sentences} &26 &16 &31\\
{Min. \# of sentences} &4 &4 &4\\ \hline

{Avg. sentence length per recipe} &12.5 &13.4 &12.5\\
{Max. sentence length} &32 &25 &19\\
{Min. sentence length} &6 &6 &7\\

\end{tabular}}
\caption{Statistics of the train, dev and test subsets of the DP dataset. Extremely short recipes with less than 4 sentences are not included into the dataset.}
\label{tab:r2vq_stats}
\end{table}

\section{Data Annotation}
\label{sec:data-annotation}

We designed a manual annotation process for creating a Dense Paraphrasing dataset. Considering the annotation complexity that comes from the coreference and interaction with hidden entities, we adopted a ``layered'' annotation schema, where a new layer of annotation is executed over the previous annotations. 
At an initial preprocessing layer, each raw recipe sentence is parsed by the Stanza pipeline \cite{Qi2020StanzaAP} for tokenization and identifying other linguistic features, including lemmatization, part-of-speech tagging, and dependencies. Given the parsed sentences, the first annotation layer includes the annotation of explicit entities and semantic roles. In this layer, annotation is done completely on the surface form of the text. The second layer includes the annotation of hidden entities, relations, and coreference (under event transformation). The third layer includes the annotation of subevents and semantic role-saturated events.

\subsection{Dense-paraphrased Event Ontology}

\begin{figure}[h]
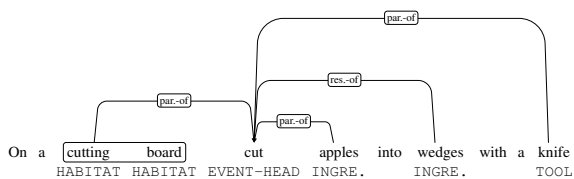

    \centering
     \resizebox{1\linewidth}{!}{
    \begin{dependency}
\begin{deptext}[column sep=.1cm, row sep=.1ex]
On \& a\& cutting \& board \& cut \& apples \& into \& wedges \& with \& a \& knife \\
\& \& \texttt{HABITAT} \& \texttt{HABITAT} \& \texttt{EVENT-HEAD} \& \texttt{INGRE.} \& \& \texttt{INGRE.} \& \& \& \texttt{TOOL} \\
\end{deptext}
\depedge{3}{5}{par.-of}
\depedge{8}{5}{res.-of}
\depedge{6}{5}{par.-of}
\depedge{11}{5}{par.-of}
\wordgroup{1}{3}{4}{a0}
\end{dependency}}
    \caption{Event annotation with explicit entities only.}
    \label{fig:crl_example}
\end{figure}

The overall goal of our annotation is to identify dense-paraphrased events from the recipe text. We define the event ontology as a set of cooking-related entities and relations. The entity types include the \textsc{event-head}, \textsc{ingredient}, \textsc{tool} and \textsc{habitat}. The relations include \textsc{participant-of} and \textsc{result-of}. Each event has only one predicative verb (\textsc{event-head}), and all the relations within the event are linked from corresponding entities to the predicate. A simple example of an explicitly saturated event is shown in Figure \ref{fig:crl_example}. \textit{cutting board}, \textit{knife} and \textit{apples} are participants of the event verb \textit{cut}, and the ingredient \textit{wedges} is the result of the event verb.

\begin{figure}[h]
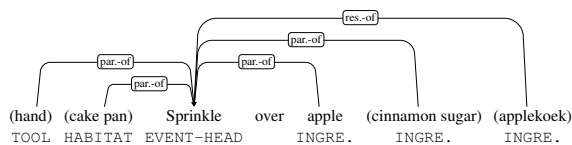

    \centering
     \resizebox{1\linewidth}{!}{
    \begin{dependency}
\begin{deptext}[column sep=.1cm, row sep=.1ex]
(hand) \& (cake pan) \& Sprinkle \& over\& apple  \& (cinnamon sugar) \& (applekoek)\\
\texttt{TOOL} \& \texttt{HABITAT} \& \texttt{EVENT-HEAD}\& \& \texttt{INGRE.}\& \texttt{INGRE.} \& \texttt{INGRE.}\\
\end{deptext}
\depedge{1}{3}{par.-of}
\depedge{2}{3}{par.-of}
\depedge{5}{3}{par.-of}
\depedge{6}{3}{par.-of}
\depedge{7}{3}{res.-of}

\end{dependency}}
    \caption{Event annotation with both explicit and hidden entities (enclosed in parenthesis).}
    \label{fig:crl_hidden_example}
\end{figure}

More importantly, for the purpose of \gls{dp}, the annotated events also involve implicitly expressed arguments by identifying their hidden entities. Consider the sentence \emph{Sprinkle over apple.} from Figure \ref{fig:crl_hidden_example}. In this event, the most plausible hidden participants and result of the event head \emph{sprinkle} are the \textsc{tool} \emph{hand}, the \textsc{habitat} \emph{cake pan} and the \textsc{ingredient}s \emph{cinnamon sugar} and \textsc{applekoek}, which can be inferred from the full recipe in Table \ref{tab:question_example}, but are not explicitly stated. The identified hidden entities should be either inferred elsewhere explicitly on the document level, or inferred based on  commonsense knowledge.

\begin{figure}[h!]
  \centering
  \includegraphics[width=1\linewidth]{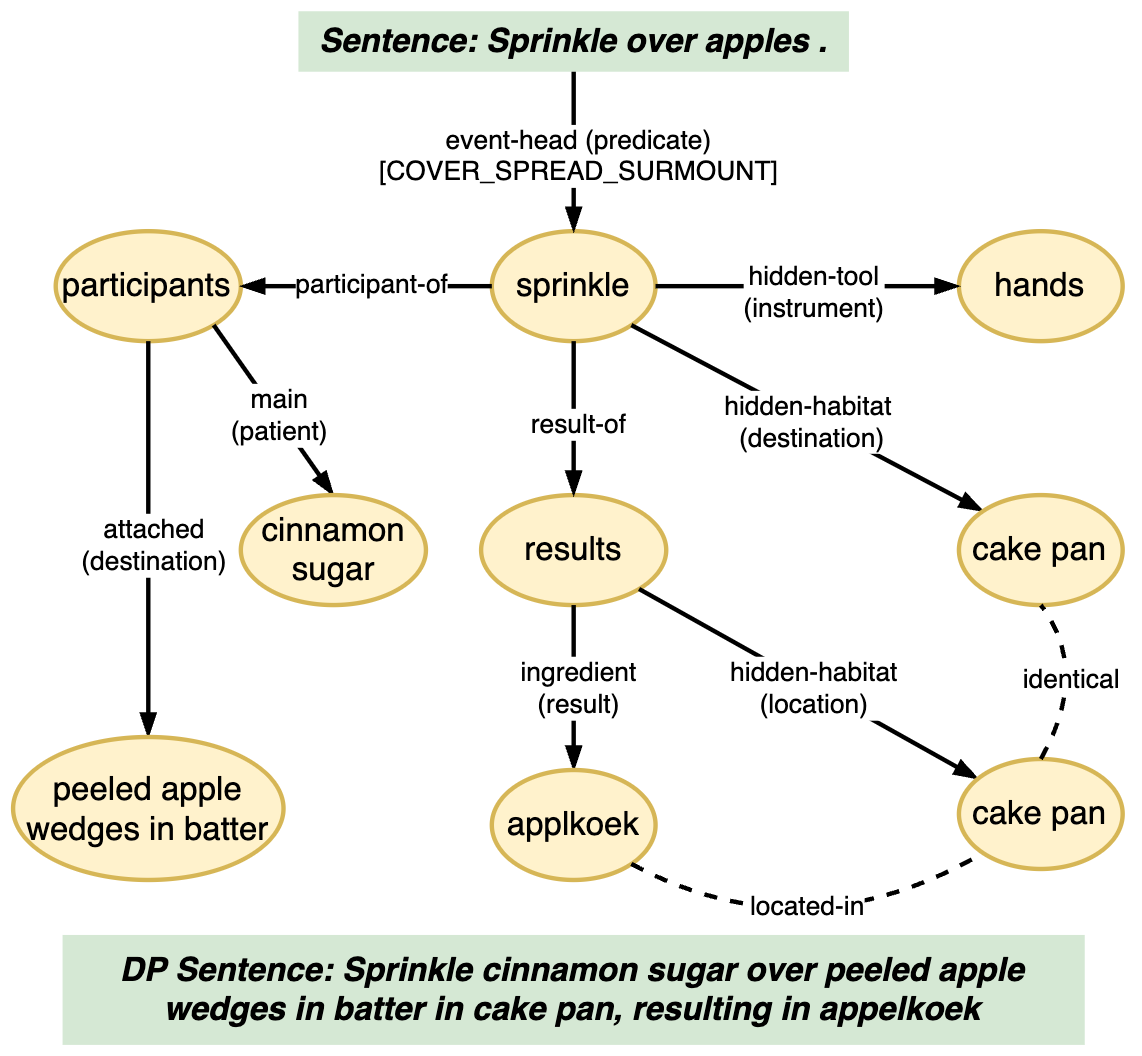}
  \caption{Graph representation of the dense-paraphrased event. Labels on the edges show the relation identified between the nodes, as well as the aligned semantic role that is enclosed. Dashed edges represent relations that can be inferred.}
  \label{fig:event_graph}
\end{figure}

The events are further enriched through subevent transformation and saturated by semantic roles.
Figure \ref{fig:event_graph} shows a complete annotated dense-paraphrased event represented as a graph. 
Each event is composed of semantic role-saturated cooking entities, relation links and event head. In the event, ingredient participants are further categorized into main (ingredients) and attached (ingredients) based on their different semantic roles. Ingredient results link the ingredients that could be transformed in this event, and also track the location change of the ingredients through a sequence of other events. In addition, more relations can be inferred from the event graph, providing further benefits to the text enrichment.
The event graph also shows the \gls{dp} of the source text \emph{Sprinkle over apples}.

\subsection{Annotating Explicit Entities and Roles}

We start the first layer of annotation by labeling explicit entities semi-automatically using a separately trained \gls{ner} model. 
Specifically, we train the Flair \gls{ner} model \cite{akbik2019flair} on another 100 recipes annotated with cooking-related entities only.\footnote{\url{https://github.com/flairNLP/flair}} The model takes a tokenized sentence and outputs the entity tag for each token in BIO format. We apply the trained \gls{ner} model to all the recipes in the dataset to generate the entities for the further validation by annotators.

To identify explicit semantic roles, we run the \gls{srl} parser from \cite{conia-navigli-2020-bridging} to label each recipe sentence. Subsequently, we ask annotators to validate and correct both frames and argument labels. We chose VerbAtlas\footnote{\url{http://verbatlas.org/}} \cite{di-fabio-etal-2019-verbatlas} as our inventory of semantic roles given its high coverage in terms of verbal lexicon, the informativeness of its human-readable roles, and its mapping to the PropBank frame inventory \cite{palmer2005proposition}, and to the BabelNet multilingual knowledge base \cite{navigli2012babelnet}. Table \ref{tab:srl_example} shows a sample recipe sentence with \gls{srl} annotation. The predicate \textit{cut} has the frame ``CUT'' in row 1; \textit{the broccoli} and \textit{into flowerets} are annotated as the patient and result of the predicate, respectively.

\begin{table}[h]
\centering
\resizebox{0.8\linewidth}{!}{
\begin{tabular}{llcl}
1 & Cut       & CUT & \texttt{B-Predicate}       \\
2 & the       & \_  & \texttt{B-Patient} \\
3 & broccoli  & \_  & \texttt{I-Patient} \\
4 & into      & \_  & \texttt{B-Result}  \\
5 & flowerets & \_  & \texttt{I-Result}  \\
6 & .         & \_  & \_     \\ 
\end{tabular}}
\caption{\gls{srl} annotation.}
\label{tab:srl_example}
\end{table}

\subsection{Annotating Hidden Entities and
Coreference}

\begin{figure*}[]
\begin{center}
\frame{
\includegraphics[scale=0.24]{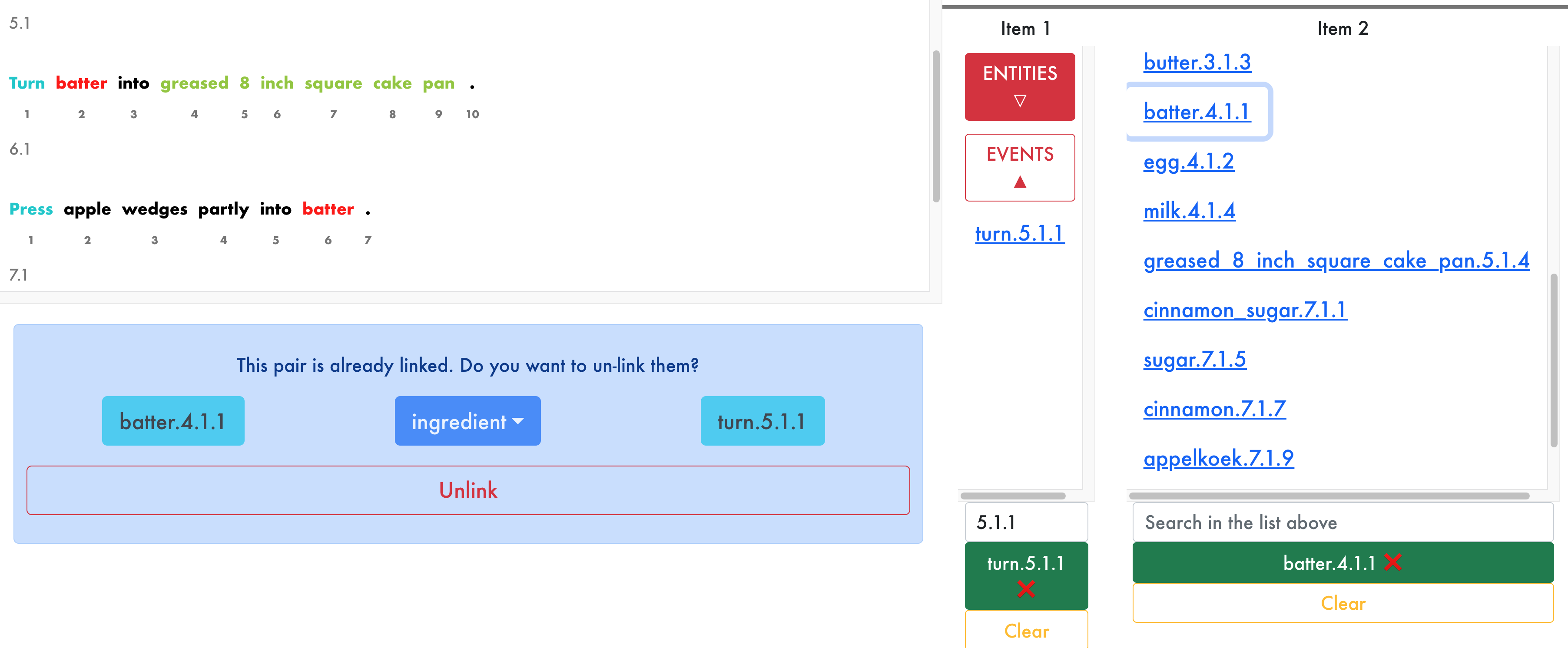}}
\caption{\gls{deep} environment for annotation of adding, linking, and coreferencing of entities.}
\label{fig:deep}
\end{center}
\end{figure*}

The second annotation layer is to annotate the events with hidden entities and relations, and link events with coreference chains on the multi-sentence level.
For this purpose, we develop \gls{deep}, a specialized annotation environment to manually annotate event structures with hidden arguments (Figure \ref{fig:deep}). 

The three main tasks in this annotation layer are: (1) adding hidden arguments; (2) linking arguments to events; and (3) linking coreferential entities. 
Annotators start from documents with surface mention spans that have been identified from the previous annotation layer, and all \gls{deep} annotation is done at document-level, namely, annotators can create long distance links within a given document. 

The \gls{deep} provides an intuitive and easy interface for pairwise linking annotation, as well as a holistic view of the document-level context using color coding of tokens related to the selected events or entities. 
Particularly, \gls{deep} provides multi-sentence linking operations between entities to include an entity in a coreference chain.
The same linking interface is also used to create role links within each event. 
This means, when there is a hidden argument from the current event, annotators do not have to first add its text surface to the entity pool and then create a link between that and its participating event or its precedents in the coreference chain, as done in a previous work \cite{msamr}. 
Instead, annotators can directly draw an entity from a previous sentence to the event that it participates in as a hidden argument. 
This one-step interface reduced annotators' workload and made the process less error-prone. 
\gls{deep} also provides an interface to add shadow entities with a free-text identifier and immediately link it to an event. 

\begin{table}[h]
\centering
\resizebox{\linewidth}{!}{
\begin{tabular}{l|rr|rr|rr}
                               & \multicolumn{2}{c|}{Train} & \multicolumn{2}{c|}{Dev} & \multicolumn{2}{c}{Test} \\ \hline
Avg. \# of entities per recipe & Exp.      & Hidden      & Exp.     & Hidden    & Exp.     & Hidden     \\
\textsc{event-head}                          &       14.0      &   N/A          &      13.6       &      N/A     &      13.3       & N/A           \\
\textsc{tool}                           &             0.6&             2.1&             0.7&           2.2&             0.6&            2.0\\
\textsc{habitat}                        &             2.8&             4.8&             2.5&           6.2&             2.5&           4.0\\
\textsc{ingredient} (participant)                     &     13.0        &   6.9          &     14.0        &     10.8      &       12.5      &   8.6         \\
\textsc{ingredient} (result)                         &       0.2      &             1.5&     0.2        &      1.4     &       0.3      &            1.7

\end{tabular}}
\caption{Statistics of cooking entity annotation from the DP dataset.}
\label{tab:crl_stats}
\end{table}

\begin{table}[]
\centering
\resizebox{\linewidth}{!}{
\begin{tabular}{l|rrr}
      Avg. \# of coref. chains per recipe    & Train & Dev & Test \\ \hline
\textsc{tool}     & 2.4           &      2.6     &        2.3          \\
\textsc{habitat}  & 5.9          &     5.8       &             5.4     \\
\textsc{ingredient}    & 16.6         &   18.2          &          16.2        \\ \hline
\textsc{ALL (explicit)} & 20.0          &    20.7        &           19.0        \\
\textsc{ALL}       & 25.0            &     26.5    &          23.9        
\end{tabular}}
\caption{Statistics of coreference chain annotation from the DP dataset. \textsc{ALL (explicit)} shows the number of chains that are composed of explicit entities \textit{only}, while \textsc{ALL} shows the number of chains that can have both explicit and hidden entities.}
\label{tab:coref_chain_count}
\end{table}

Table \ref{tab:crl_stats} shows the average number of annotated entities per recipe. The average number of event verbs (14) is much great than the average recipe length (8) from Table \ref{tab:r2vq_stats}, indicating that many recipe sentences tend to involve more than one event. Ingredient participants are the most prevalent entity type under both explicit and hidden settings. Recipes also have more hidden ingredient results, tools and habitats instead of explicit ones, showing the importance of hidden arguments for understanding cooking recipes or instructional texts, in general. 

Table \ref{tab:coref_chain_count} shows the average number of coreference chains per recipe for different types of entities as well as the explicit entities. Similarly, the chains of ingredients still dominant in numbers, followed by the chains of habitats and tools.

\subsection{Assembling Dense-paraphrased Events}

The last annotation layer is to enrich an event with any subevent transformation and semantic role saturation.
For the subevent structure, we have defined two types of event end state, namely {\sc transformation} and {\sc location-change} (\textsection\ref{sssec:subevent}). 
To identify the end state type of each event, we collected 208 unique verb senses that are assigned by the \gls{srl} parser to our data, and hand-split those into three categories: \textsc{transformation}, \textsc{location-change} or neither. Then we assign the end state type based on the category of the event verb sense. Table \ref{tab:verb_sense} shows some examples from each category.

\begin{table}[h]
\small
        \centering
        \resizebox{\linewidth}{!}{
        \begin{tabular}{lll}
        \hline
             \textsc{Sense} & \textsc{Verbs} & \textsc{Category} \\
              \hline
            {CONVERT} & {melt, cream, evaporate} & Transformation\\
            \hline
            {SPILL\_POUR} & {pour, ladle, drip} & Loc. Change\\
            \hline
            {AMELIORATE} & {enhance, improve, round out} & N/A\\
            \hline
        \end{tabular}}    

        \caption{Example event verbs and their senses along with the assigned end state type.}
        \label{tab:verb_sense}
    \end{table}

We incorporate the subevent structure by replacing the entity with its transformed version.
Specifically, for entities of result ingredient from \textsc{transformation} events, 
we paraphrase it into a form of \textbf{Ee} + \textsl{NP}
(e.g., \emph{minced garlic}, \emph{heated water}). 
For \texttt{loc.change} events, the entities remain the same, but an additional habitat is added to mark the destination of the entities as shown in Figure \ref{fig:event_graph}. This paraphrase is also ``chainable'' through the coreference. For example. if the entity \textit{onions} has been transformed into \textit{chopped onions} from the previous event, the paraphrase of the entity from the next event would be \textit{saut\'eed chopped onions}.

The last step is saturating events with semantic roles. Some semantic roles can be naturally linked to explicit entities, e.g., the \textit{broccoli} from Table \ref{tab:srl_example} can be labeled as both Patient and \textsc{INGREDIENT}.
Specifically, for any given sentence, we align the annotation of entities and roles that share the same event predicate (marked as \texttt{Predicate} in \gls{srl} and \texttt{EVENT-HEAD} in cooking entities). 
Semantic roles that are \emph{claimed} are merged with corresponding cooking entities.\footnote{If the text span of a cooking entity C1 is overlapped with that of a semantic role S1, we describe it as C1 claims S1.} 
For example, in the sentence \emph{Transfer peas to the saucepan quickly.}, the role \emph{to the saucepan}\textsubscript{\texttt{[destination]}} will be merged with the entity \emph{saucepan}\textsubscript{\texttt{[HABITAT]}}, and the text span \emph{peas} is both the entity \texttt{INGREDIENT} and the role \texttt{Theme}. 
The un-overlapped roles such as the \texttt{Attribute} \textit{quickly} will be categorized as modifiers to this cooking event. 
Each hidden entity by default is also assigned with its most plausible semantic role and further validated manually.\footnote{Role-entity mapping: \texttt{Instrument} to \texttt{TOOL}, \texttt{Location/Destination} to \texttt{HABITAT}, \texttt{(Co-)Theme/Patient} to \texttt{INGREDIENT} (participant) and \texttt{Result} to \texttt{INGREDIENT} (result).}

\subsection{Inter-annotator Agreement}
\label{sec:crl_iaa}

Among all the annotation layers, entities and coreference chains have  relied heavily on manual annotation and validation. Thus, to measure the annotation quality of that, we double-annotated a subset of 50 randomly selected recipes and computed the \gls{iaa} on the subset. 
We first measure the agreement on cooking entities using Cohen's $\kappa$ \cite{Cohen1960ACO}, and Table \ref{tab:iaa_entity} shows the agreement. The \textsc{explicit} roles annotation results in a higher Cohen's $\kappa$ of 0.90. As a comparison, \textsc{hidden} role annotations have a lower score of 0.61, indicating the difficulty of the task to recover the hidden information from the text. By looking at the annotation, we find that for \textsc{hidden} entities, the disagreement primarily comes from the missing annotation of ingredients and the different annotation of tools when the context information is insufficient. For example, given the sentence \textit{whisk the eggs.}, all the plausible hidden tools such as \textit{whisk}, \textit{fork} or \textit{chopsticks} can be added by annotators.
Overall, the annotation results in a $\kappa$ of 0.73, showing the reliability of the cooking entity annotation.

\begin{table}[h]
\centering
\resizebox{1\linewidth}{!}{
\begin{tabular}{l|rr}
    Entity type& \multicolumn{1}{l}{\# of entities from \textsc{Ann1}/\textsc{Ann2}} & \multicolumn{1}{l}{Cohen's $\kappa$} \\ \hline
\textsc{explicit} & 707 / 693                      & 0.900                  \\
\textsc{hidden} & 1006 / 979                      & 0.608                  \\\hline
\textsc{ALL} & 1713 / 1672                     & 0.730                 
\end{tabular}}
\caption{Inter-annotator agreement on cooking entities measured by Cohen's $\kappa$.}
\label{tab:iaa_entity}
\end{table}

We further measure the agreement on the coreference of cooking entities on the same subset of 50 recipes. Following previous work on coreference evaluation \cite{pradhan-etal-2012-conll,cattan-etal-2021-realistic}, we use the CoNLL-2012 F1 score that consists of three coreference evaluation metrics to measure the \gls{iaa}. Since we also cover hidden entities in \gls{dp}, the coreference chains may consist of both hidden and explicit entities.
Table \ref{tab:iaa_coref} shows the agreement scores on the coreference annotation. The CoNLL-F1 scores on the chains of each entity type is balanced, ranging from 56 to 58. For the overall agreement score, the chains of explicit entities only result in a higher F1 score (85.71), comparing to the chains of both explicit and hidden entities (57.46). Similar to the entity annotation agreement, this is expected, as annotating hidden entities tends to be more challenging to annotators. 

We also compare the \gls{iaa} with other popular coreference datasets (Table \ref{tab:iaa_coref_compare}). The CoNLL-2012 shared task \cite{pradhan-etal-2012-conll} annotates the coreference of explicit entities and events in OntoNotes \cite{ontonotes}, and only reports the MUC score \cite{Vilain1995AMC} as the \gls{iaa} metric. Textual-based NP enrichment (TNE) annotates the coreference of NPs and reports both CoNLL-F1 and MUC.\footnote{TNE reports a CoNLL-F1 of 79.8. It is not included in the table for the comparison with the CoNLL-2012 dataset.}
Both RED \cite{ogorman-etal-2016-richer} and MS-AMR \cite{ogorman-etal-2018-amr} annotate event coreference on the multi-sentence level, but MS-AMR designs the annotation schema over the AMR architecture. Compared to CoNLL-12 and TNE, our explicit chain agreement results in a higher MUC score (87.64). This may be due to the fact that the surface form of references in the cooking domain has less variance, e.g., no pronouns, thus making it easier to achieve higher agreement. Our overall agreement on coreference chains (57.46) is lower than the RED and MS-AMR. One reason is the \gls{iaa} for both RED and MS-AMR is computed on two subsets of documents with identical entity/event annotation, so unlike the \gls{dp} annotation, the potential disagreement on the entities will not cascade to coreference annotation. Another reason might be the larger number of hidden coreferences in our data, which increases the difficulty of the task. In general, our \gls{iaa} on coreference is in the rough range of the agreement scores from existing works, demonstrating the reliability of the coreference annotation in \gls{dp}.

\begin{table}[h]
\centering
\resizebox{\linewidth}{!}{
\begin{tabular}{l|rr}
      Entity type    & \multicolumn{1}{l}{\# of chains from \textsc{Ann1}/\textsc{Ann2}} & \multicolumn{1}{l}{CoNLL-F1} \\ \hline
\textsc{tool}     & 178 / 172                      & 58.08                  \\
\textsc{habitat}  & 319 / 310                      & 56.35                  \\
\textsc{ingredient}    & 783 / 785                      & 58.27                  \\ \hline
\textsc{ALL (explicit)} & 955 / 970                      & 85.71                  \\
\textsc{ALL}       & 1280 / 1267                     & 57.46                 
\end{tabular}}
\caption{Inter-annotator agreement on coreference chains measured by CoNLL-F1.}
\label{tab:iaa_coref}
\end{table}

\begin{table}[h]
\centering
\resizebox{\linewidth}{!}{
\begin{tabular}{l|ll}
\hline
      Chain type    & \multicolumn{1}{l}{Dataset} & \multicolumn{1}{l}{MUC} \\ \hline
\textsc{explicit}     & CoNLL-2012 \cite{pradhan-etal-2012-conll}   & 83.03                  \\
     & TNE \cite{elazar2021text}   & 83.60                  \\
  & \gls{dp} (This work)                      & 87.64                  \\
    \hline
    \hline
      Chain type    & \multicolumn{1}{l}{Dataset} & \multicolumn{1}{l}{CoNLL-F1} \\ \hline
\textsc{ALL} & RED \cite{ogorman-etal-2016-richer}                      & 65.50                  \\
       & MS-AMR \cite{ogorman-etal-2018-amr}                     & 69.90 \\
    & \gls{dp} (This work)                    & 57.46  
\end{tabular}}
\caption{Inter-annotator agreement on coreference chains from various datasets.}
\label{tab:iaa_coref_compare}
\end{table}

\subsection{Annotators}
We posted annotation positions within several University-wide distribution lists, available to all students within the various departments targeted. We hired 10 student annotators for the recipe annotation work. They were paid at the University-mandated rate of \$15/hour for student research assistants. All annotators were students at a US-based university, ranging from undergraduate to master's program. 

The complete annotation task includes: (1) the validation of the explicit entities and semantic roles produced by the trained models; (2) the annotation of relations, hidden entities and coreference; 3) the annotation and validation of subevent verbs and transformed entities. All the annotators are trained to be familiarized with the annotation guideline and annotation examples before they start the task.

\label{ssec:dataprep}
\begin{figure}[h] 
\resizebox{\columnwidth}{!}{%
\begin{tikzpicture}[node distance=1.5cm, align=center]

\tikzstyle{base} = [rectangle, draw=black, minimum width=1cm, minimum height=1cm,
text centered, font=\sffamily, fill=white,
]
\tikzstyle{process} = [base, fill=orange!15, font=\ttfamily]
\tikzstyle{multidocument} = [base, shape=tape, draw=black, fill=white, tape bend top=none, double copy shadow]
\tikzstyle{predproc} = [base, shape=predproc]
\tikzstyle{manual} = [base, shape=trapezium, trapezium angle=110, trapezium stretches=true, fill=blue!30]
\tikzstyle{final} = [base, shape=trapezium, trapezium left angle=70, trapezium right angle=110, trapezium stretches=true]

  \node (source_data)   [multidocument]              {Source Data \\ (Recipes)};
  \node (selection)     [predproc, right of=source_data, xshift=3cm]          {Document \\ Selection \& \\ Preprocessing};
  \node (ner)           [process, below of=source_data, yshift=-0.5cm]   {Custom-trained \\ \gls{ner}};
  \node (srl)           [process, below of=selection, yshift=-1cm]   {Automatic  \\ SRL parser};
  \node (ner_correction)[manual, below of=ner] {Human \\ Correction};
  \node (srl_correction)[manual, below of=srl] {Human \\ Correction};
  \node (deep)          [manual, below of=ner_correction] {\gls{dp} annotation \\ (\gls{deep}) };
  \node (assembly)         [process, below of=deep, right of=deep, xshift=1cm]  {Event Assembly};
  \node (final)         [final, below of=assembly]  {Final Dataset};
  \path (source_data.south) -- coordinate (aux1) (source_data.south|-ner.north);
  \path (deep.south) -- coordinate (aux2) (deep.south|-assembly.north);
  \path (assembly.south) -- coordinate (aux3) (assembly.south|-final.north);
  
  \draw[->]             (source_data) -- (selection);
  \draw[dashed]  ([xshift=-1cm]ner.north west|-aux1) -- ([xshift=1cm]srl.north east|-aux1);    
  \draw[->]             (selection) -- (ner);
  \draw[->]             (selection) -- (srl);
  \draw[->]             (ner) -- (ner_correction);
  \draw[->]             (srl) -- (srl_correction);
  \draw[->]             (ner_correction) -- (deep);
  \draw[->]             (deep) -- (assembly);
  \draw[dashed]  ([xshift=-1cm]ner.south west|-aux2) -- ([xshift=1cm]srl.north east|-aux2);    
  \draw[->]             (srl_correction) -- (assembly);
  \draw[dashed]  ([xshift=-1cm]ner.north west|-aux3) -- ([xshift=1cm]srl.north east|-aux3);    
  \draw[->]             (assembly) -- (final);
\end{tikzpicture}
}
  \caption{Annotation Process, \gls{dp} annotation process on the left-hand side, and \gls{srl} on the right-hand side. Dotted lines indicate difference layers of the phased annotation.}
\label{fig:annotationflow}
\end{figure}
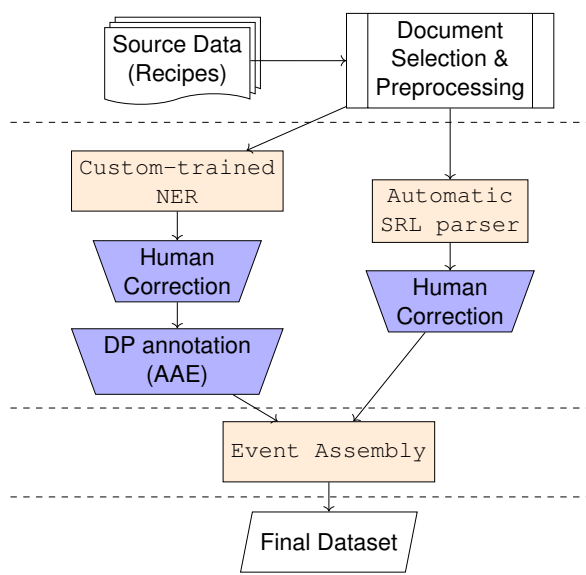

\section{Experiments}
\label{sec:experiments}
We conduct experiments to explore the utility of the \gls{dp} method and provide further insights about our annotated datasets. In \textsection\ref{ssec:qa} we first show the utility of the \gls{dp} for rich text inference by conducting an experiment with a \gls{qa} task that lets models answer a set of questions. We then compare results from two \gls{qa} models; one trained with original text, and another trained with human-readable \gls{dped} text. In \textsection\ref{ssec:crl_model}, we show the reusability of dense-paraphrased events by training models that can generate paraphrases in machine-readable format. In \textsection\ref{ssec:crl_transfer}, we explore and discuss the transferability of dense-paraphrased events. We show a case study by letting the fine-tuned text generation models infer over out-of-domain procedural texts.

\subsection{Question Answering on DP Questions}
\label{ssec:qa}

\subsubsection{Generating Questions}
\label{sssec:qg}
To prepare the data for the \gls{qa} task, we start by generating question-answer pairs from our annotated data.
Automatic question generation is the task of generating questions and answers from a given text. Recent QG work includes rule-based approaches \cite{Lindberg2013GeneratingNL,Labutov2015DeepQW}, the combination of templates and semantic relations \cite{dhole2021synqg,pyatkin2021asking}, and end-to-end neural generation models \cite{Du2017LearningTA,Yuan2021ImprovingNQ}. For our purpose, we adopt a template-based method to generate questions that are able to solicit new information from DP-enriched text. 

We first propose text templates for each type of question. Then we generate \gls{qa} pairs by populating the templates with slots in a cloze test style. 
Table \ref{tab:question_type} shows the text templates for three types questions that we use for the \gls{qa} task.
\textsc{Elision} identifies arguments (ingredients in most cases) that are omitted from a text, but can be understood from context. 
\textsc{Implicit} covers both hidden tools and habitats introduced in the text. This is distinct from \textsc{elision}, as these are not solved merely through contextual clues. 
\textsc{Object Lifespan} covers the transformation of the ingredients through multiple events. For example, the answer to the question \emph{What is in the appelkoek?} contains all distinct ingredient participants of the events that result in either the referent entities of \emph{applkoek} or itself.


Candidates slots from the text templates (colored spans in Table \ref{tab:question_type}) are acquired from \gls{dped} events we created earlier. We also set the constraints to only keep the events with at least one hidden cooking entity for template population, so that the generated questions need to be answers by inferring from the context, rather than purely ``memorizing'' original text spans.

After a text template is populated, it is further processed to improve the readability of the generated question. We change word inflections and insert articles and agreements. For the templates with [habitat\_phrase] and [tool\_phrase] slots, we fill those with corresponding \texttt{LOCATION} or \texttt{INSTRUMENT} spans from \gls{srl}. If a slot is filled with a hidden entity, we run a BERT-based model \cite{BERT} to get the most likely preposition given the sentence as  context through the masked language modeling task. Modifiers are populated in the same order as they were in the original sentence.

To increase the variety of questions, we allow adjunct slots in the text templates. As shown in Table \ref{tab:question_type}, adjunct slots include tool/habitat phrases and modifiers. For example, 
one \textsc{Elision} question can be as short as \emph{What should be saut\'{e}ed?} or \emph{\ldots saut\'{e}ed in the saucepan with the spatula until browned?} with all the adjunct slots. 
We argue it is helpful to generate questions more challenging to the systems. Adding more adjunct slots completes the context for the question, but also introduces unseen context if the slots contain hidden entities.



\begin{table*}
\small
        \centering
        \resizebox{1\linewidth}{!}{
        \begin{tabular}{| m{0.15\linewidth} | m{0.4\linewidth} | m{0.4\linewidth} | }
        \hline
             \textsc{Question Type} & \textsc{Text Template} & \textsc{Question-Answer Pair}\\
             \hline
             \multirow{1}{*}{Elision} & What should be \textcolor{brown}{verb} \textcolor{red}{[habitat\_phrase]} \textcolor{blue}{[tool\_phrase]} \textcolor{green}{[modifiers]}? --- \textcolor{purple}{ingredient\_obj} & {What should be \textcolor{brown}{cut} \textcolor{red}{on the board} \textcolor{blue}{with a knife} \textcolor{green}{into eighths}? --- \textcolor{purple}{peeled apples}} \\
             \hline
             \multirow{2}{*}{Implicit} & What do you use to \textcolor{brown}{verb} \textcolor{cyan}{obj} \textcolor{red}{[habitat\_phrase]} \textcolor{green}{[modifiers]}? --- \textcolor{blue}{tool} & {What do you use to \textcolor{brown}{saut\'e} \textcolor{cyan}{the chopped onions} \textcolor{red}{[in the pan]}? --- \textcolor{blue}{spatula}} \\
             & Where do you \textcolor{brown}{verb} \textcolor{cyan}{obj}  \textcolor{blue}{[tool\_phrase]} \textcolor{green}{[modifiers]}? --- \textcolor{red}{habitat\_phrase} & {Where do you \textcolor{brown}{arrange} \textcolor{cyan}{the slices}  \textcolor{green}{[into rounds]}? --- \textcolor{red}{in the casserole}} \\ \hline
                 \multirow{2}{*}{Obj. Lifespan} & What is in \textcolor{cyan}{obj}? --- \textcolor{purple}{ingredient\_objs} & {What is in the \textcolor{cyan}{appelkoek}? --- \textcolor{purple}{apples, batter and cinnamon sugar}} \\
             & How did you get \textcolor{cyan}{ingreObj}? --- \textcolor{brown}{event\_phrase} & {How did you get the \textcolor{cyan}{appelkoek}? --- \textcolor{brown}{by sprinking cinnamon sugar over peeled apple wedges with batter} } \\
             \hline
        \end{tabular}}   
        \caption{Text templates and example of generated questions. The squared brackets ([...]) in the templates indicates adjunct slots.}
        \label{tab:question_type}
    \end{table*}
    
\subsubsection{Results}
\label{ssec:results}

We first apply the aforementioned question generation method to both the training and validation sets from our datasets. Table \ref{tab:question_count} shows the number of questions we are able to generate from each subset.
We fine-tune the T5 text generation model \cite{2020t5} to perform a \gls{qa} task on the questions that are generated from the training set, and evaluate on the validation set using exact match (EM) and token-level F1 score (F1) following \citet{squad2}.

\begin{table}[]
\centering
\small
\resizebox{0.7\linewidth}{!}{
\begin{tabular}{l|rr}
      Question Type    & Train & Dev  \\ \hline
\textsc{Elision}     & 2,251           &      340            \\
\textsc{Implicit}  & 1,385         &     261          \\
\textsc{Obj. Lifespan}    & 2,392         &   312              \\ \hline
\textsc{ALL}       & 6,028            &     913        
\end{tabular}}
\caption{Number of generated questions from each subset.}
\label{tab:question_count}
\end{table}

For the model training, We format each input instance to {\small \texttt{"question: \{question\_str\} context: \{recipe\_str\}"}} that includes the question string and the raw text of the whole recipe as context regardless of the question scope or implicity. We use this model setup as our baseline.
As a comparasion, to understand whether our annotation can help answer dense-paraphrased questions, we fine-tune another T5 model on the same question set, but we replace the context string with its human readable paraphrase. For example, given a piece of the context: {\small \texttt{saut\'e until browned}}, we change it to {\small\texttt{\{using a spatula\} \{on the cutting board\}, saut\'e \{chopped onions\} until browned \{resulting in saut\'eed onions\}}} to recover the hidden objects. Similar to the question generation, we run a BERT-based model to get the most likely preposition of each phrase component.

Table \ref{tab:qa_res} shows the \gls{qa} results from the models on the \gls{dp} questions per type. Comparing to \textsc{Elision}, the \textsc{Base} model performs better for \textsc{Implicit} on EM (12.22 up), but slightly worse on F1 (1.22 down). By examining the examples from the model output, we find that the correct \textsc{Elision} answer tends to include multiple ingredients, while the \textsc{Implicit} answer is a single tool or habitat. Since part of the ingredients from the \textsc{Elision} answer are not hidden from the text, the model is able to extract those easily thus resulting in a higher F1 on \textsc{Elision} from the \textsc{Base} model. The long answer for \textsc{Elision} also makes the model difficult to predict a complete answer, so the EM score is much lower on such questions. \textsc{Obj. Lifespan} is the most challenging question type to the models as it leverages the inference over coreference from multiple events. Thus the scores are the lowest among all three question types. The big gap between the EM and F1 for \textsc{Obj. Lifespan} also indicates the limitations of the models to identify hidden referent entities from a multi-sentence context.

Comparing to the \textsc{Base} model, \textsc{Base+DP} perform better on every question type (Table \ref{tab:qa_res}). The improvement from incorporating \gls{dp} is larger on \textsc{Implicit} (23.37/17.73 up on EM/F1). Through analysis, we find that the \textsc{Base+DP} model improves \textsc{Implicit} by recovering the hidden tools and habitats that are implicated or completing the entity text form (e.g. \emph{spoon} to \emph{\textbf{large wooden} spoon}). For \textsc{Elision} and \textsc{Obj. Lifespan}, the model can improve it primarily by completing the subevent structure of the entities (e.g. \emph{apples} to \emph{\textbf{peeled} apples}) in the answer, and make coreference chains more explicit and identifiable.
Overall, the gain on scores shows the utility of \gls{dp} information on the \gls{qa} task.

\begin{table}[]
\centering
\small
\resizebox{1\linewidth}{!}{
\begin{tabular}{l|rr|rr:r}
             & \multicolumn{2}{c|}{\textsc{Base}} &
            \multicolumn{2}{c:}{\textsc{Base+DP}}\\ 
          & EM      & F1  & EM      & F1   & Count \\ \hline

\textsc{Elision}  &     35.29 &    58.47   & \textbf{41.47} &  \textbf{60.04}  &   340 \\
\textsc{Implicit}    &  47.51 &     57.25    &  \textbf{70.88} &  \textbf{74.98}  & 261 \\ 
\textsc{Obj. Lifespan}    &  18.48 &     50.12    &  \textbf{27.16} &  \textbf{59.82}  & 312 \\
\hline
\textsc{All}   &    33.04  &   55.27    &   \textbf{44.99}  & \textbf{64.24}  &  913     \\ \hline
\end{tabular}}
\caption{\gls{qa} results from the T5 model fine-tuned on recipe text only (\textsc{Base}) and recipe text with \gls{dp} (\textsc{Base+DP}).}
\label{tab:qa_res}
\end{table}

\subsection{In-Domain \glsentrylong{dp} Modeling}
\label{ssec:crl_model}
We further explore the possibility to train models that can infer \gls{dp} automatically. In the previous section, we have shown that common textual inference task like \gls{qa} can benefit from DP text, so it is critical to evaluate the generalizability and tranferability of the \gls{dp}.

\begin{figure}[h!]
  \centering
  \includegraphics[width=1\linewidth]{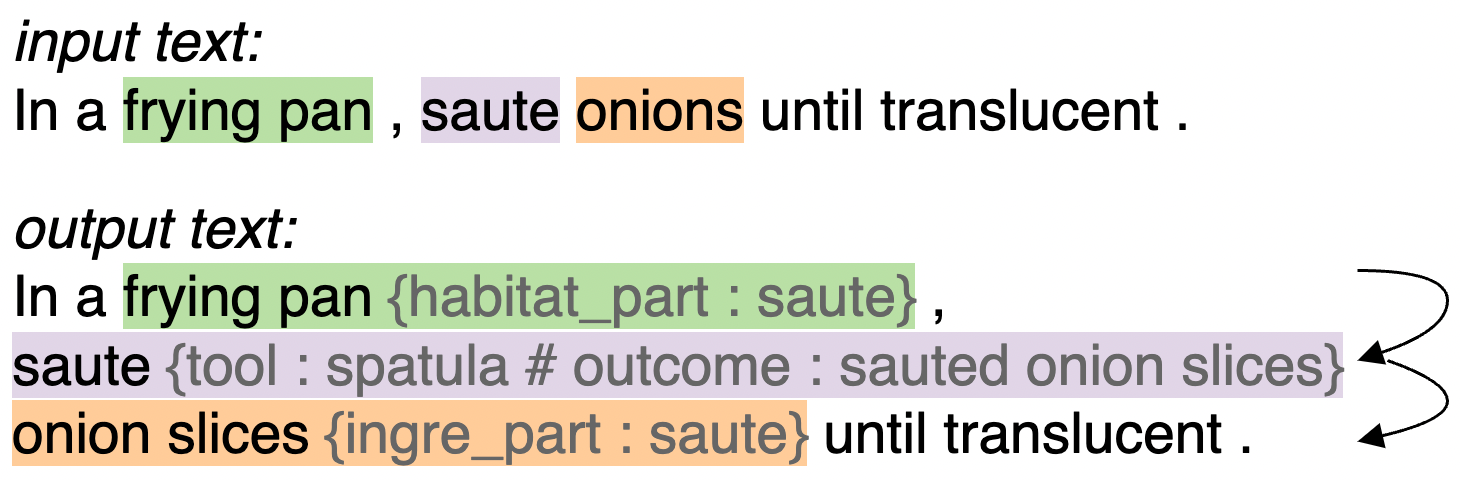}
  \caption{Example of T5 model input and output for \gls{dp} generation task. Each cooking role is wrapped by a pair of curly brackets (\{...\}). Cooking roles at the same position are separated by hashtags (\#).}
  \label{fig:crl_modeling_input}
\end{figure}

It is a common practice to apply pretrained large language model as an end-to-end system for text generation tasks such as machine translation and text summarization, and recent advance in prompt learning has shown that additional text prompts or indicators embedded in the model input can effectively improve the performance on various tasks \cite{Liu2021PretrainPA}. Inspired by that, we formalize the DP modeling as a text generation task for the baseline. Specifically, we apply the T5 text generation model on the raw recipe text and let it generate output text in the machine readable format. 
Figure \ref{fig:crl_modeling_input} illustrates an example of model input and output. Each cooking role is represented as similar to a key-value pair that is enclosed by ``\{...\}''. The key-value pair positioned after an explicit entity (e.g. \emph{frying pan}) stores the relation and the event head it is linked to, while the key-value pair after the event head (e.g. \emph{saute}) stores the hidden entities separated by ``\#''. Transformed ingredients are edited in place in the output text (e.g. \emph{onions} to \emph{onion slices}).
In practice, each model input instance is the concatenation of maximum three consecutive sentences until the model input length is reached.
The model output text is the machine readable paraphrase of the input text. Compared to the human readable paraphrase from the previous \gls{qa} task, this output format is more structured and has the flexibility to be converted to human-readable paraphrase or event graphs. In addition, similar to the prompt-based training paradigm, the extra symbols and text structure built by curly brackets and hashtags can be considered as indicators or prompts that regulate the models to perform better.  

We fine-tune a T5-based model to perform the \gls{dp} generation task on the training set, and test on the validation set. We use precision, recall and F-score as the evaluation metrics for the task.
For the first experiment, we evaluate the model output only based on the cooking role, regardless of its value. 
Table \ref{tab:crl_modeling_res} shows the model results on different cooking entity sets. Comparing to the hidden entity set, the model performs better on the explicit set (83.79 on \textsc{All}), indicating the ability of the model to identify the relations between explicit entities and the event head. The model is also able to retrieve the majority of the hidden cooking roles (65.21 on \textsc{All}), showing the feasibility to apply \gls{dp} on the new text. Among all types of the entities, the model attains the lowest F1 score on ingredient results (33.48 from \textsc{Hidden}, 26.67 from \textsc{Explicit}). It is expected due to the facts that entities from this set have the least number (1.6 per recipe on average), and the ingredient results tend to be more diverse in the recipe text.

To further explore how well the model can generate complete \gls{dp} from scratch, we also evaluate the model output based on the exact match of the cooking role and its value. Table \ref{tab:crl_modeling_res_strict} shows the results of this experiment. Comparing to the explicit set, the F1 score on the hidden set drops significantly (81.08 down to 28.04). It shows that the inference and reasoning over all the hidden text remains a very challenging task to current large language models. For our data specifically, the higher ratio of the hidden entities and the entity variance from the subevent transformation makes it a challenging task to the model. Attempts to improve the results may include multi-task learning to generate cooking roles and values separately, and iterative training to utilize the data more efficiently. We will leave it to future discussion.

\begin{table}[h]
\small
\centering
\resizebox{1\linewidth}{!}{
\begin{tabular}{l|rrr|rrr}
            & \multicolumn{3}{c|}{\textsc{Hidden}} & \multicolumn{3}{c}{\textsc{Explicit}} \\ 
         Entity type  & P      & R     & F1    & P      & R     & F1   \\ \hline
\textsc{tool} &54.11 &57.08 &55.56 & 75.86& 77.19& 76.52 \\
\textsc{habitat} &63.25 &64.51 &63.88 &80.63 &94.44 &86.99 \\
\textsc{ingredient} (participant) &78.65 &73.68 & 76.09& 88.84& 95.17 & 91.89 \\
\textsc{ingredient} (result)          &41.76 & 27.94 & 33.48 & 28.57& 25.00& 26.67  \\ \hline
\textsc{All}  & 66.53 & 64.20 & 65.21 & 80.65& 87.25& 83.79 \\ \hline
\end{tabular}}
\caption{Results of the \gls{dp} generation task based on the match of the entity type.}
\label{tab:crl_modeling_res}
\end{table}

\begin{table}[h]
\small
\centering
\resizebox{1\linewidth}{!}{
\begin{tabular}{l|rrr|rrr}
            & \multicolumn{3}{c|}{\textsc{Hidden}} & \multicolumn{3}{c}{\textsc{Explicit}} \\ 
        Entity type   & P      & R     & F1    & P      & R     & F1   \\ \hline
\textsc{tool} &40.26	&42.47	&41.33 & 68.97	&70.18&	69.57 \\
\textsc{habitat} &29.11	&29.68	&29.39 &76.28&	89.35&	82.3 \\
\textsc{ingredient} (participant) &26.48&	24.81&	25.62& 86.47&	92.63&	89.44 \\
\textsc{ingredient} (result)          &20.88&	13.97&	16.74 & 28.57& 25.00& 26.67  \\ \hline
\textsc{All}  & 28.85 & 28.09 & 28.40 & 78.05& 84.41& 81.08 \\ \hline
\end{tabular}}
\caption{Results of the \gls{dp} generation task based on the exact match of both the entity type and entity value.}
\label{tab:crl_modeling_res_strict}
\end{table}

\begin{figure*}[h!]
  \centering
  \includegraphics[width=1\linewidth]{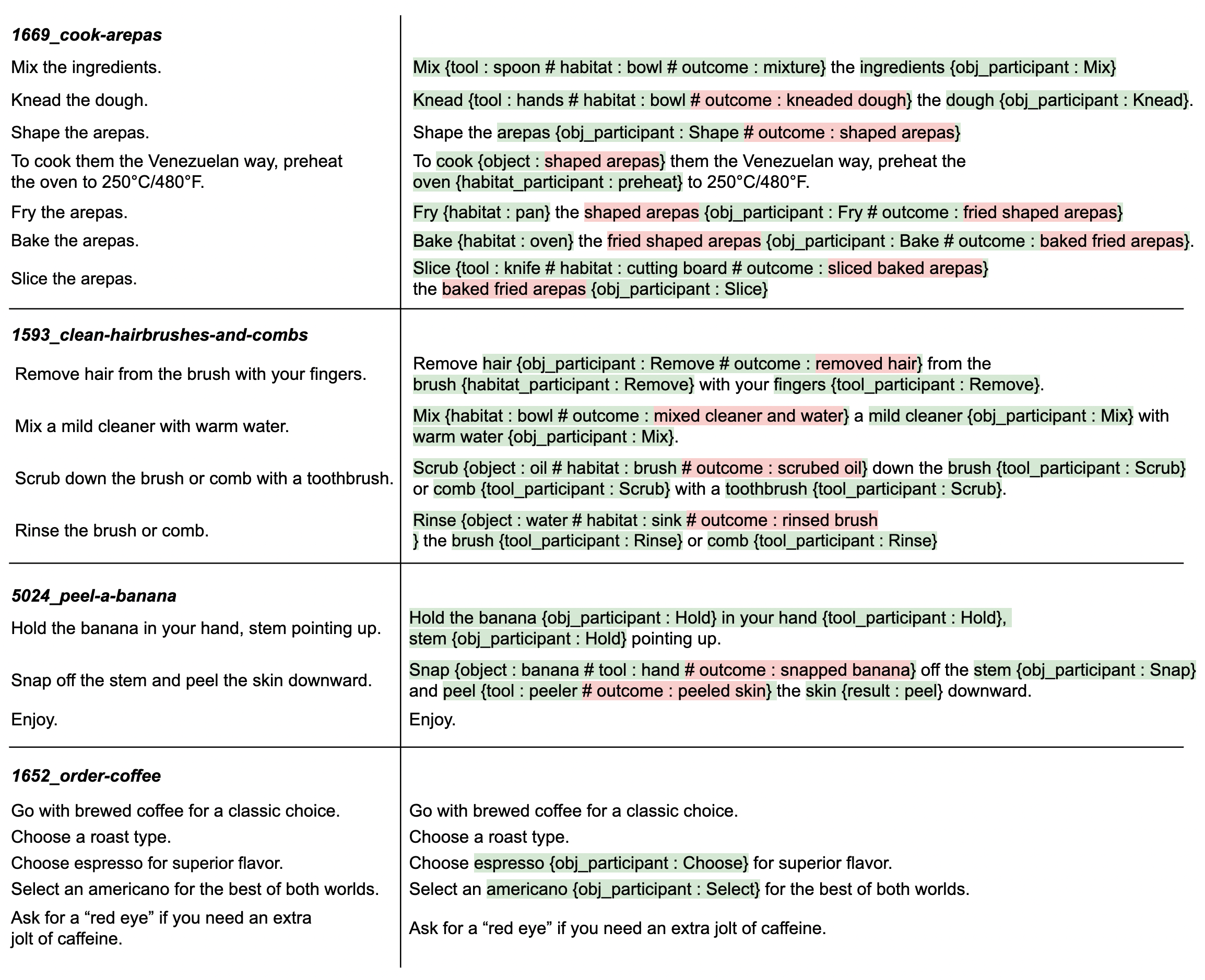}
  \caption{Automatic \gls{dp} annotations on WikiHow articles from different domains. The first column shows the article title and step sentences; the second column shows the model output in machine-readable format. Green spans mark the cooking roles and relations generated by the T5 model; red spans mark the roles that are generated through subevent transformation.}
  \label{fig:crl_transfer}
\end{figure*}

\subsection{Out-of-Domain DP Modeling}
\label{ssec:crl_transfer}

While our proposed datasets focus on the sub-domain of cooking recipes, it is critical to show that the DP strategy and datasets can be adapted to raw data in the same style but out of the domain under a transfer learning scenario. As a subset of procedural texts, cooking recipes share a lot of commons with other types of procedural texts. For example, they all tends to be imperative and instructional comparing to descriptive texts like news articles, and certain syntactic components and semantic roles are naturally omitted from sentences, making them an effective transfer learning resources. 

Based on this intuition, we show a case study of the results by applying the T5 DP generation model that is fine-tuned on our training set to WikiHow articles.\footnote{\url{https://www.wikihow.com/}}
For this experiment, we use the articles from the WikiHow corpus curated by \citep{zhang-etal-2020-reasoning} that is originally for the goal-step inference tasks. Specifically, we pick four articles from different domains as the test data.
For each article we use the main body of the text that consists of several headline sentences as the steps.
The model training and model input/output is similar to the process described in \textsection\ref{ssec:crl_model} and depicted in Figure \ref{fig:crl_modeling_input}. Since the data is not confined to ``cooking'' in this experiment, we replace any occurrence of the key \textit{ingredient} in the model input with \textit{object} (e.g. \textit{ingre\_part} $\rightarrow$ \textit{object\_part}). For the subevent-transformed ingredient object, we keep maximum two end states for readability, e.g., \textit{sauted minced peeled garlic} will be shortened to \textit{sauted minced garlic}.

The generation results on the four unseen WikiHow articles are shown in Figure \ref{fig:crl_transfer}. The first article is an in-domain cooking recipe from a different data source; the other three articles are instructions for activities from different domains. 
By looking at the results, our model performs better on the first two articles, even though they have longer text and contain more context information. The first article is an in-domain recipe, and the model is able to recover all the relations and most of the hidden entities correctly. For ingredient like \textit{arepas} that appear in multiple sentences, the subevent transformation shows its state change from \textit{shaped arepas} to \textit{sliced baked arepas} through the process.
The results on the second article shows the effectiveness of our datasets and model being applied to out-of-domain data. Our defined DP event structure can be naturally transferred to text with clear steps and intermediate goals (e.g. \textit{Mix a mild cleaner with warm water}). The model could mispredict the actual values of the hidden entities due to the limitations from the domain-specific vocabulary inventory. For example, in the second article, the predicted hidden entity is \textit{oil} from the sentence \textit{``Scrub down the brush ...''}. The subevent transformation strategy however, is able to complement the result without the limitation of the vocabulary. For example, in the second sentence, the hidden result ingredient of the event \textit{mix} is \textit{mixed cleaner and water}. Similarly in the last sentence, we are able to generate \textit{rinsed brush} that carries the subevent state effectively.\footnote{The tool \textit{brush} has the role \texttt{PATIENT} from SRL, so it can be subevent-transformed.}

Comparing to the first two, the last two articles seem to be more challenging to the model. Although the text is short, the third article involves rather complex spatial actions (e.g. \textit{snap off, peel downward, etc}) that may confuse the model. The part-whole relations of entities (e.g. \textit{banana v.s. skin v.s. stem}) can also lead to semantically ambiguous subevent paraphrases such as \textit{snapped stem / banana, peeled skin / banana}.
The last article is different from the others in the sense that it has a less clear step-goal structure and the events are not actions interacting with physical objects. These differences make the proposed moethod less effective on texts of this type.
In general, the case study shows the usefulness of the DP strategy and the dataset we created under a transfer learning scenario. The machine-readable DP paraphrase of procedural texts can be used as resources for downstream text inference tasks, and the generation of DP paraphrase itself can be formalized as a text enrichment task. In the future, we will expand the \gls{dp} evaluation on general procedural texts so that a quantitative study can be conducted.

\subsection{Technical Details}
\label{tech_details}
We adopt the pretrained T5 text generation model \cite{2020t5} as the base model, and fine-tune it with different task goals to perform all the experiments. 
For the QA task, we fine-tune \textsc{T5-Base} model for 10 epoches on 4 NVIDIA Titan Xp GPUs at each experiment run. For the \gls{dp} generation task, we fine-tune \textsc{T5-Base} model for 15 epoches at each experiment run.
It took roughly 1-2 hours to finish the training of each run. We adapt the training script from \url{https://huggingface.co/valhalla/t5-base-qa-qg-hl}.

\section{Conclusion}
\label{sec:conclusion}
In this paper we define  {\it \glsentryfull{dp}}, the task of enriching a text fragment (lexeme, phrase, or sentence) such that contextual ambiguities are eliminated, contextual anchors or variables are supplied, and any implied arguments are made textually explicit. We describe a corpus of recipes annotated with both hidden arguments and coreference chains, as well as \gls{srl}. We outlined our \gls{dp} procedure that creates an enriched textual dataset, that can then be used to train transformer-based models for text inference tasks. We reported experiments with a T5 model for answering a set of questions about cooking, constructed from the corpus. The results of these experiments show that \gls{dp} significantly improves performance on the three types of dense-paraphrased questions used for evaluation, {\sc Elision}, {\sc Implicit} and {\sc Obj. Lifespan} by recovering the hidden entities that are implicated and completing the subevent structure of the entities. We also reported the experiments for generating \gls{dp} automatically under both in-domain and out-of-domain settings. The results show the feasibility of modeling \gls{dp} and the challenges it poses to current large language models.

We believe that \gls{dp} has potential to help in a broad range of \gls{nlp} applications, using both human-readable and machine-readable paraphrases. In particular, applications and tasks involving abstractive inferencing can benefit from the  dynamic tracking and decontextualized redescriptions of entities appearing in a coreference chain. The notion of following an entity as it changes through a developing narrative or text can be computationally encoded using the technique described here, giving rise to a history or biographical model of an entity.  
We hope to extend the \gls{dp} procedure to include creating vector representations of \gls{dp} that can be fit into a broader range of computational models. We also intend to include reference to the ``vertical typing'' of an expression (type inheritance) from online resources with definitional texts, such as Wikipedia or WordNet (e.g., onion $\in$ vegetable, poodles $\in$ dogs). This would further enhance the utility of the resulting \gls{dped} data for logical inference tasks.



\bibliography{dpte}
\bibliographystyle{acl_natbib}

\end{document}